\documentclass[lettersize,journal]{IEEEtran}
\usepackage{amsmath,amsfonts}
\usepackage{algorithmic}
\usepackage{algorithm}
\usepackage{array}
\usepackage[caption=false,font=normalsize,labelfont=sf,textfont=sf]{subfig}
\usepackage{textcomp}
\usepackage{stfloats}
\usepackage{url}
\usepackage{verbatim}
\usepackage{graphicx}
\usepackage{cite}
\usepackage{xcolor}
\usepackage[hidelinks]{hyperref}
\hypersetup{
  pdfauthor={}, pdftitle={}, pdfsubject={}, pdfcreator={}, pdfproducer={}
}

\begin{document}
\title{Memory-Efficient 2D/3D Shape Assembly of Robot Swarms}

\author{Shuoyu Yue, Pengpeng Li, Yang Xu, Kunrui Ze, Xingjian Long, Huazi Cao, Guibin Sun

\thanks{This work was supported in part by the National Natural Science Foundation of China under Grants 62503028, 62373019 and 625B2017, and in part by the Natural Science Foundation of Hangzhou under Grant 2025SZRJJ0152. \textit{(Corresponding authors: Kunrui Ze; Guibin Sun.)}

Shuoyu Yue is with the Cavendish Laboratory, University of Cambridge, Cambridge CB3 0HE, United Kingdom (e-mail: shuoyu.yue@gmail.com). This work was conducted during his research internship at Beihang University.

Pengpeng Li is with the School of Artificial Intelligence, Beihang University, Beijing 100191, China (e-mail: pengpengli@buaa.edu.cn).

Yang Xu, Kunrui Ze, Xingjian Long, and Guibin Sun are with the School of Automation Science and Electrical Engineering, Beihang University, Beijing 100191, China (e-mail: \{xuyangx, kr\_ze, longxj, sunguibinx\}@buaa.edu.cn).

Huazi Cao is with the College of Artificial Intelligence, Zhejiang University, Hangzhou 310058, China (e-mail: caohuazi@wioe.westlake.edu.cn).}
}



\maketitle

\begin{abstract}
Mean-shift-based approaches have recently emerged as a representative class of methods for robot swarm shape assembly. They rely on image-based target-shape representations to compute local density gradients and perform mean-shift exploration, which constitute their core mechanism. However, such representations incur substantial memory overhead, especially for high-resolution or 3D shapes. To address this limitation, we propose a memory-efficient tree representation that hierarchically encodes user-specified shapes in both 2D and 3D. Based on this representation, we design a behavior-based distributed controller for assignment-free shape assembly. Comparative 2D and 3D simulations against a state-of-the-art mean-shift algorithm show one to two orders of magnitude lower memory usage and two to four times faster shape entry. Physical experiments with 6 to 7 UAVs further validate real-world practicality.
\end{abstract}


\section{Introduction}
\label{sec:intro}
\IEEEPARstart{I}{n} nature, insect groups can assemble into functional structures through purely local interactions~\cite{mccreery2022hysteresis}. For instance, fire ants dynamically build living bridges to span gaps and transport resources, with stability emerging from simple individual rules. Inspired by such collective behaviors, swarm robotics is extensively explored as a means for large groups of robots to self-organize into desired 2D or 3D configurations, enabling applications from cooperative construction to search-and-rescue in complex environments~\cite{
nedjah2025swarm,
nitti2025collective,
zhang2022aerial
}.

Existing swarm shape assembly approaches can be categorized into assignment-based and assignment-free paradigms. Assignment-based methods dynamically allocate each robot to a unique target location, and then drive each robot to its assigned target~\cite{
shan2024distributed,
shorinwa2023distributed,
sakurama2020multi,
wang2020shape
}. Some work also constructs and reconfigures multi-level coordination structures online to support the allocation and execution process~\cite{zhu2024self,zhang2023self}. However, conflicts can arise from concurrent local decisions and may require additional mechanisms such as task swapping to ensure consistency~\cite{wang2020shape}. Closely related, traditional formation control research typically assumes a predefined goal assignment and focuses on accurate, smooth, and collision-free tracking of a prescribed formation~\cite{
zhao2019bearing,
liu2023distributed,
grafe2025dmpc,
pham2024safety,
rai2024safe}. While these formation control methods achieve strong performance in small-scale multi-robot systems, their scalability is often limited.

Assignment-free shape assembly approaches are also extensively studied~\cite{slavkov2018morphogenesis,sabattini2011arbitrarily,alhafnawi2021self}. One class of assignment-free methods encodes the desired shape as an image-like map that each robot queries locally to generate motion toward the target shape. This class is often referred to as image-based methods, ranging from Kilobot-based works~\cite{
rubenstein2014programmable,
march2024honeybee,
pluhacek2025decentralised} to mean-shift exploration strategies~\cite{sun2023mean,zhang2024mean,lu2025concurrent}, as well as recent branches focusing on image moments~\cite{liu2024self} and continuous sculpting~\cite{curtis2024continuous}. Among them, mean-shift-based methods are the most closely related to our approach. These methods are promising, as a representative mean-shift-based method has been shown to outperform both assignment-based and assignment-free approaches in 2D~\cite{sun2023mean}. However, the reliance on full-grid image maps makes memory usage grow exponentially with dimensionality, thereby rendering direct extensions to 3D substantially more memory-intensive and potentially prohibitive for resource-constrained robots.

\begin{figure}[t]
\centering
\includegraphics[width=1.0\linewidth]{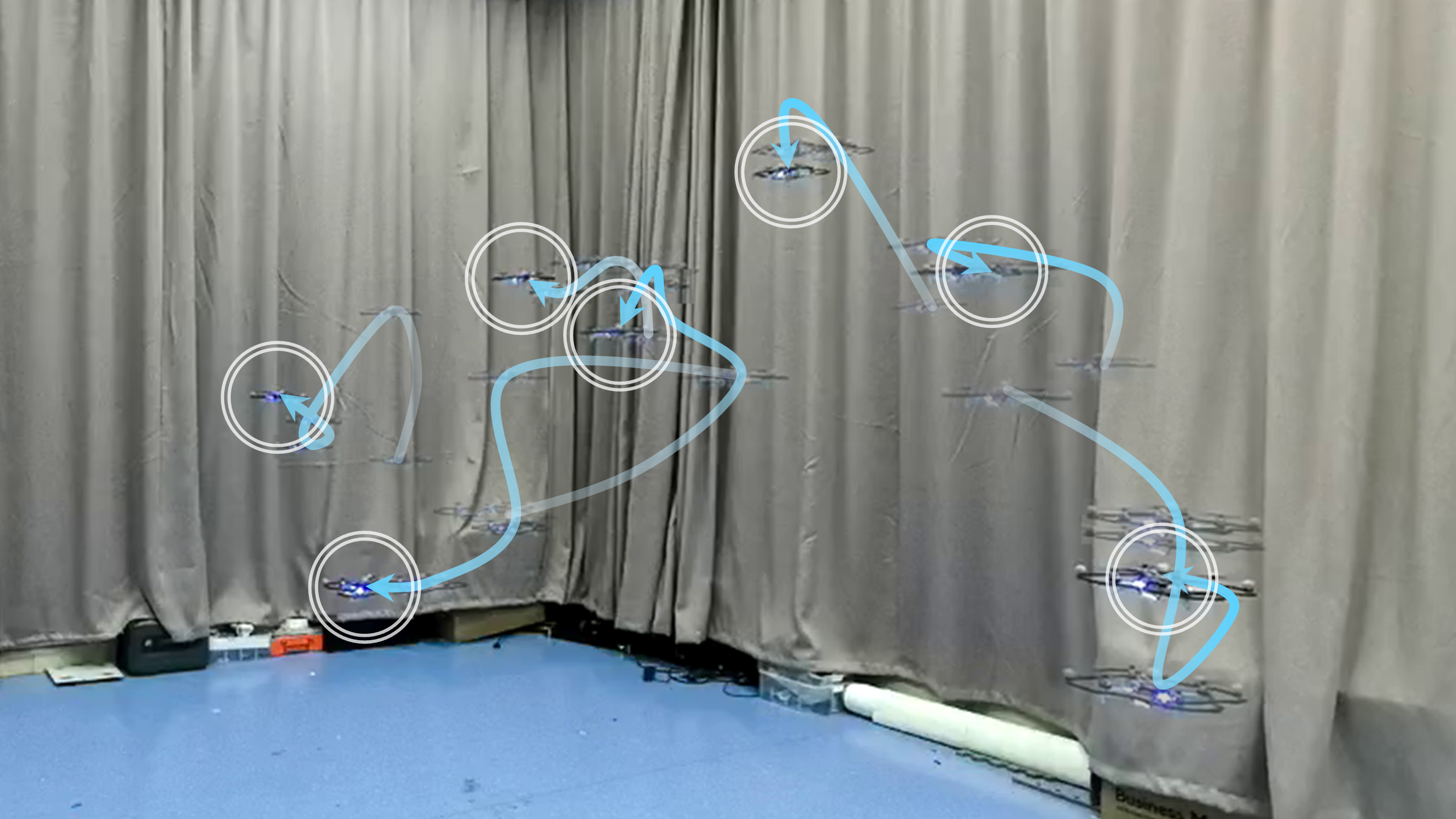}
\caption{Physical experiment with 7 UAVs forming a 3D arrow. Final UAV positions are marked with white circles. Trajectories are shown as smooth blue curves passing through five sampled positions for each UAV, with color gradually deepening with time. Note that the curves depict smoothed connections rather than exact flight paths.}
\label{fig:front}
\end{figure}

Whether assignment-based or assignment-free, most existing methods remain confined to 2D, while scalable 3D extensions are rare and technically demanding. For aerial swarms, Morgan et al. proposed a distributed auction algorithm with trajectory planning~\cite{morgan2016swarm}, but it incurs substantial memory and communication costs because each robot must store all target points and iteratively exchange bids. Underwater bio-inspired swarms can exhibit 3D patterns via visual signaling~\cite{berlinger2021implicit}, yet these behaviors are coarse and transient rather than precisely defined shapes. Robotic 3D construction has also been studied~\cite{
cabral2023design,
swissler2020fireant3d,
zhao2025modular,
zhao2025softsnap}. Such approaches typically build structures sequentially from fixed bases or seed robots, unlike the free-space shape assembly studied in this work.

In this letter, we propose a memory-efficient, assignment-free shape assembly framework for large-scale robot swarms in both 2D and 3D. We encode user-specified shapes into a tree map (quadtree in 2D and octree in 3D) to reduce memory usage, and design a behavior-based distributed controller based on this map. Extensive 2D and 3D simulations demonstrate reliable formation of complex shapes. Compared with a baseline mean-shift method~\cite{sun2023mean}, our approach reduces memory usage by one to two orders of magnitude and improves shape entry by a factor of two to four. Only one baseline is considered in this paper because our primary goal is to address the memory bottleneck in extending~\cite{sun2023mean} from 2D to 3D. UAV experiments further confirm practical feasibility.

\section{Methodology}
\label{sec:metho}
\begin{figure*}[t]
\centering
\includegraphics[width=1.0\textwidth]{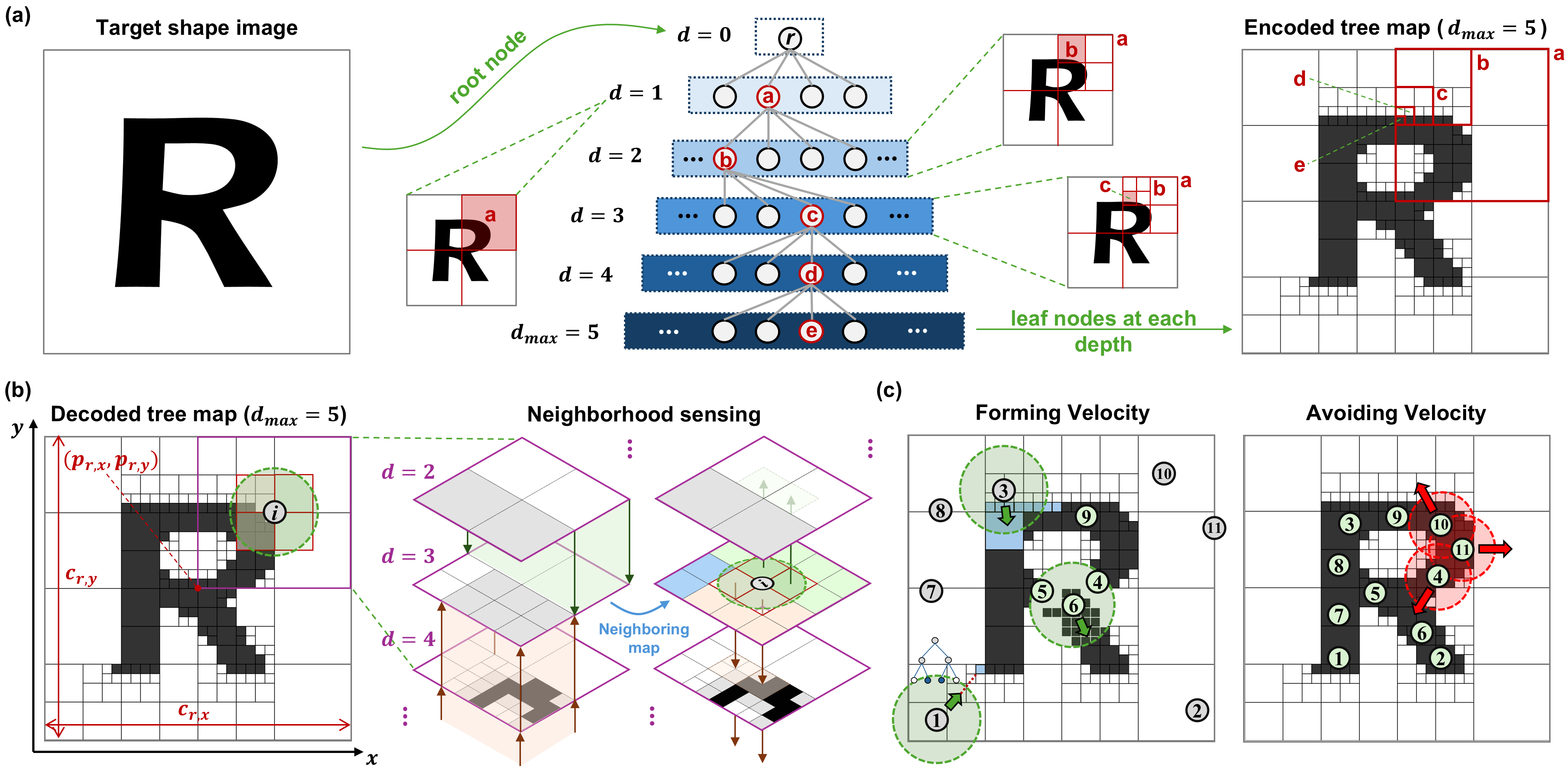}
\caption{Overview of the Proposed Framework. (a) Shape encoding: The target shape image is shown on the left, while the middle and right panels depict the encoded tree map as a tree structure and a grid view, respectively. A complete encoding path from node $a$ at $d=1$ to node $e$ at $d_\text{max} = 5$ is highlighted in red in both the tree structure and the grid view. (b) Shape decoding: Physical lengths are assigned to the root node $r$, and thus to the entire tree map. A neighboring map is established at $d=3$ and used for each robot $i$ to perceive its neighborhood. The sensing range of robot $i$ is shown as a green circle and the sensed neighborhood is outlined in red. (c) Shape assembly controller: Examples of the two velocity commands under different cases are illustrated. The collision-avoidance range of each robot is shown as a red circle.}
\label{fig:overview}
\end{figure*}

In this section, we address the challenge of using target shape images as maps for swarm shape assembly with low memory usage, especially in 3D. We present a tree-based encoding-decoding method that converts a dense image into a hierarchical tree map, and design a distributed, assignment-free controller with forming and collision-avoidance velocities. An overview is shown in Fig.~\ref{fig:overview}.

\subsection{Shape Encoding}
\label{subsec:shapeEncoding}
The target shape encoding is shown in Fig.~\ref{fig:overview}(a), where a user-specified binary image is converted into a tree map. For clarity, all figures in this section use 2D examples. Since the method applies to both 2D and 3D, we use $\lambda$ to denote the dimensionality. The maximum tree depth $d_\text{max}$ controls map resolution and memory cost, with $d_\text{max}=5$ in Fig.~\ref{fig:overview}.

Encoding is performed via breadth-first subdivision. A root node $r$, representing the entire image, is first created at depth $d=0$ and then subdivided into $n_{\text{div}}=2^{\lambda}$ equal child nodes, which form layer~1 ($d = 1$) of the tree. For each node in this layer, if all its covered pixels share the same color (black or white), it is marked as a leaf node; otherwise, it is further subdivided. This process continues layer by layer until $d=d_\text{max}$. Any mixed-color node at depth $d_\text{max}$ is assigned black if the majority of its pixels are black, and white otherwise. Fig.~\ref{fig:overview}(a) shows an example: node $a$ at layer~1 contains pixels of different colors, so it is subdivided into four child nodes, including node~$b$ at layer~2. Then node~$b$ is further divided into four child nodes, including node~$c$ at layer~3.

At depth $d_\text{max}$, redundant storage can occur when a parent’s children are all leaf nodes of the same color. We remove this redundancy via a post-processing merge: whenever all children of a parent are leaf nodes with identical color, we delete the children and convert the parent into a leaf node. This process repeats until no further merges are possible. As shown in Fig.~\ref{fig:overview}(a), the resulting tree map is defined by its leaf nodes, which form heterogeneous cells with binary colors $\rho \in \{0,1\}$. The memory usage $m_\text{tree}$ of a tree map is computed as
\begin{equation}
\label{eqn:memory}
\begin{gathered}
m_\text{tree} = 
n_\text{div} m_\text{link} + 
n_\text{middle} (n_\text{div} + 1) m_\text{link} + \\
n_\text{leaf} (m_\text{link}+m_\text{color}).
\end{gathered}
\end{equation}

Here, the first term accounts for the root node, while $n_\text{middle}$ and $n_\text{leaf}$ denote the numbers of middle-layer and leaf nodes, respectively. The parameter $m_\text{link}$ represents the memory required for a unidirectional link between a parent and a child node. Each link is stored as an integer node index occupying 4 bytes, which is sufficient for all 2D and 3D shapes and resolutions considered in this study. The parameter $m_\text{color}$ corresponds to the memory occupied by a single color value. Since the color values in a tree map are binary, $m_\text{color}$ is set to one bit. By contrast, the method in~\cite{sun2023mean} employs a full-grid map, where a grayscale gradient field is constructed from the original image to guide robots into the target shape. The memory required for a full-grid map is
\begin{equation}
\label{eqn:memory-cmp}
m_\text{full-grid} = n_\text{cell} \times m_\text{color},
\end{equation}
where $n_\text{cell}$ is the total number of cells and $m_\text{color}$ is 4 bytes for a single-precision floating-point number since $\rho \in [0,1]$ in this case.

\subsection{Shape Decoding and Robot Localization}
\label{subsec:shapeDecoding}

In decoding, each node $n$ in the tree is associated with physical quantities, namely its center position $\mathbf{p}_n \in \mathbb{R}^\lambda$ and size $\mathbf{c}_n \in \mathbb{R}^\lambda$. An example for the root node $r$ is shown in Fig.~\ref{fig:overview}(b) (left). This is achieved by establishing a global coordinate frame and redefining the pixel size in the target shape image. In the swarm shape-assembly literature, it is sometimes assumed that each robot has access to a global map, a shared reference frame, and its own position in that frame~\cite{wang2020shape,sun2023mean}. Since we extend the method in~\cite{sun2023mean}, we maintain this assumption. Note that we do not assume all-to-all communication: each robot can access only the positions of robots within its sensing range $r_\text{sense}$, and we denote this neighbor set by $\mathcal{N}^\text{robot}_i$. Although a global map is provided in our setting, we use the term ``sensing'' to refer to an algorithmic query based on local information, rather than onboard perception.

Since robots are intended to occupy the black region, we define the pixel size as
\begin{equation*}
c_\text{pixel} = \sqrt[\lambda]{\frac{s_r \times n_\text{robot}}{n_\text{black}}},
\end{equation*}
where $n_\text{robot}$ and $n_\text{black}$ denote the numbers of robots and black pixels, respectively. The term $s_r$ represents the space occupied by each robot, approximated in 2D by a circle of radius $r_\text{avoid}/2$, where $r_\text{avoid}$ is the collision-avoidance distance. In 3D, to account for UAV downwash effects in our physical experiments, this circle is generalized to an ellipsoid elongated along the z-direction~\cite{preiss2017downwash}; the same compensation is applied in the collision-avoidance velocity in Eq.~(\ref{eqn:v_avoid}). We also define the attraction $\beta_n$ of each node $n$, which measures the fraction of black region within its coverage. It is computed bottom-up: for a leaf node, $\beta_n$ equals its color, and for a parent node, $\beta_n$ is the average attraction of its children. Thus, $\beta_n \in [0,1]$, with larger values indicating stronger attraction.

Once the tree map is physically decoded, each robot $i$ can be localized within the map. Two types of localization are considered depending on the target. The first associates robot $i$ with a leaf node $l_i$. This is performed through a top-down search: starting from the root node $r$, the robot's position $\mathbf{p}_i$ is compared with the center $\mathbf{p}_r$ to select the corresponding child node at depth $d=1$. This procedure is repeated at each depth until a leaf node is reached. The second type localizes robot $i$ at a specified depth. The procedure is the same as the first one but terminates once the target depth is reached. If the target depth exceeds the actual depth of the leaf nodes, virtual child nodes can be temporarily generated to complete the search and released after use.

To enable shape assembly control, each robot $i$ must sense its neighborhood in the map, focusing on the black cells. This is nontrivial in a tree map because spatial adjacency is partially lost during encoding: cells that appear adjacent in the grid view may correspond to leaf nodes far apart in the tree. To recover this information, we build a full-grid neighboring map at depth $d_\text{map}$ during decoding and use it for control. This map fully overlaps the tree map at $d=d_\text{map}$ and stores all data needed for neighborhood sensing. Fig.~\ref{fig:overview}(b) illustrates this process, where only the top-right quarter is shown for clarity. We set $d_\text{map}=d_\text{max}-2$ (thus $d_\text{map}=3$ here) for reasons discussed later in this subsection.

Fig.~\ref{fig:overview}(b) (middle) visualizes the focused region of the tree map at depths $d=2,3,4$: black/white cells denote leaf nodes, and gray cells are parents that are further divided. The core idea is to construct a full grid at $d=3$, linking each cell to the corresponding leaf nodes. The neighboring map contains three types of cells: (i) overlapping leaf nodes at $d=3$, (ii) overlapping parent nodes at $d=3$, and (iii) lying inside larger leaf nodes at shallower depths. Cells of types (i) and (iii) directly link to their leaf nodes regardless of size, whereas type (ii) cells record only the overlapped parent node and traverse its subtree (two layers deep) on demand to retrieve all child leaf nodes. Linking of types (ii) and (iii) is highlighted in red and green in Fig.~\ref{fig:overview}(b).

The resulting neighboring map is shown in Fig.~\ref{fig:overview}(b) (right), where cells of types (i)-(iii) are marked in blue, red, and green, respectively. To use this map, robot $i$ is first localized to $d_\text{map}$, and the neighboring cells within its sensing range $r_\text{sense}$ (outlined in red) are identified. The linked leaf nodes are then retrieved according to cell type and returned to robot $i$, enabling it to sense surrounding leaf nodes and select the black ones for control. The choice of $d_\text{map}$ trades off memory and computation: a deeper $d_\text{map}$ creates more cells and increases map generation/storage cost, whereas a shallower $d_\text{map}$ requires deeper subtree traversals for type (ii) cells at each iteration. In the worst case, the number of visited descendants grows exponentially with traversal depth, which can noticeably increase the neighborhood lookup time and become a bottleneck in the real-time control loop. We set $d_\text{map}=d_\text{max}-2$ as a practical compromise between memory cost and efficiency.

In this work, we adopt a tree structure to represent the target shape, although more aggressively compressed representations such as shell-based models or functional descriptions could further reduce memory usage. This choice is motivated by the requirements of distributed shape assembly, where each robot must repeatedly access a fine-resolution spatial structure to compute motion commands. Higher compression typically removes local geometric details more severely, which can degrade control efficiency and stability. In contrast, our tree-based representation achieves substantial memory savings compared with a dense full grid while preserving an explicit and complete spatial description.

\subsection{Shape Assembly Controller}
\label{subsec:Control}
The behavior-based distributed controller for robot swarm shape assembly is designed based on the tree map. The control command of robot $i$ is given in the form of a velocity vector as follows:
\begin{equation*}
\mathbf{v}_i = \mathbf{v}_i^{\text{form}} + \mathbf{v}_i^{\text{avoid}},
\end{equation*}
which consists of two components: a forming velocity that drives the robot to assemble the target shape, and a collision-avoidance velocity that ensures collision avoidance. Fig.~\ref{fig:overview}(c) provides a 2D illustration of the velocity commands.

\subsubsection{Forming Velocity}
The forming velocity is designed based on node search within the tree map. The definition of the forming velocity $\mathbf{v}_i^\text{form}$ for robot $i$ is given by
\begin{equation}
\label{eqn:v_form}
\mathbf{v}_i^\text{form} = \kappa_1
\frac{\sum_{n\in\mathcal{N}_i} \omega_n (\mathbf{p}_n - \mathbf{p}_i)}
{\sum_{n\in\mathcal{N}_i} \omega_n},
\end{equation}
where $\mathbf{p}_n$ and $\mathbf{p}_i$ denote the center position of node $n$ and the current position of robot $i$, respectively. The set $\mathcal{N}_i$ contains all the candidate nodes for robot $i$, the parameter $\omega_n$ denotes the weight assigned to each node $n \in \mathcal{N}_i$, and $\kappa_1$ is a positive control gain.

During shape assembly, the role of the controller depends on the robot's relative state with respect to the target shape. When a robot is outside or far from the target region, the controller prioritizes efficient guidance into the shape. Once the robot is near or within the shape, the focus shifts toward filling the shape and achieving a uniform distribution. Accordingly, different node search strategies are applied, leading to distinct definitions of $\mathcal{N}_i$. To quantify this relative state for each robot $i$, we introduce a parameter $\eta_i$ as below:
\begin{equation}
\label{eqn:eta}
\eta_i = 
\min_{n \in \mathcal{N}_i^{\mathrm{sense}}}
\left\|
2(\mathbf{p}_n - \mathbf{p}_i) \oslash \mathbf{c}_n
\right\|_\infty,
\end{equation}
where $\oslash$ denotes an element-wise division and $\mathbf{c}_n$ is the size of node $n$. The set $\mathcal{N}_i^{\text{sense}}$ contains all black leaf nodes within distance $r_\text{sense}$ of robot $i$. It is obtained with the neighboring map and is highlighted in blue for robot 3 in Fig.~\ref{fig:overview}(c). The parameter $\eta_i$ quantifies the relative proximity of robot $i$ to the shape: the robot is inside the shape if $\eta_i \leq 1$ and outside otherwise, with $\eta_i \gg 1$ if $|\mathcal{N}_i^{\text{sense}}|=0$. Based on this parameter, the full candidate set $\mathcal{N}_i$ is defined as follows:
\begin{equation*}
\begin{gathered}
\mathcal{N}_i =
\begin{cases} 
\mathcal{N}_i^{\text{tree}} \cup \mathcal{N}_i^{\text{sense}}, & \eta_i > 1, \\ 
\mathcal{N}_i^{\text{virtual}}, & \eta_i \le 1,
\end{cases} \\
\end{gathered}
\end{equation*}
where $\mathcal{N}_i^{\text{tree}}$ and $\mathcal{N}_i^{\text{virtual}}$ are candidate sets obtained with different searching methods.

The set $\mathcal{N}_i^{\text{tree}}$ contains a leaf node identified through a tree search. Two distinct scenarios are considered for the tree search. First, if robot $i$ is inside the map but outside the target shape (e.g., robot 1 in Fig.~\ref{fig:overview}(c)), the search starts from the parent node of its current leaf node $l_i$ and proceeds downward, selecting at each depth the child node with the maximum attraction $\beta_n$, until a black leaf node is reached, which is highlighted in blue in Fig.~\ref{fig:overview}(c). Second, if robot $i$ is outside the map, the search begins from the root node and, at each depth, selects the nearest child node with $\beta_n > 0$, continuing downward until a leaf node is reached.

The set $\mathcal{N}_i^{\text{virtual}}$ consists of virtual nodes defined on an exploration layer at depth $d_\text{max}$. These virtual nodes are generated dynamically through a subdivision strategy and removed afterward, thereby incurring no additional memory overhead. For each robot $i$, localization is performed at depth $d_\text{max}$, and a uniform grid of virtual nodes is generated within its sensing range $r_\text{sense}$. The color of each virtual node is then determined by checking whether it lies within a black leaf node in $\mathcal{N}_i^{\text{sense}}$. Then the virtual nodes occupied by each neighboring robot $j \in \mathcal{N}_i^\text{robot}$ are excluded, and the remaining black virtual nodes constitute $\mathcal{N}_i^\text{virtual}$. Here, a virtual node is considered occupied if its distance to any neighboring robot is less than $r_\text{avoid}/2$. An example of $\mathcal{N}_i^{\text{virtual}}$ for robot 6 is illustrated in Fig.~\ref{fig:overview}(c).

\begin{algorithm}[t]
\caption{\textbf{Compute Forming Velocity}}
\label{alg:v_form}
\begin{algorithmic}[1]
\STATE \textbf{Input:} 
$T$: Tree map, $P$: Robot positions
\STATE \textbf{if} robot $i$ is outside $T$ \textbf{then}
\STATE \hspace{0.5cm} Select $\mathcal{N}^\text{tree}_i$ by tree search
\STATE \hspace{0.5cm} $\mathcal{N}_i \gets \mathcal{N}^\text{tree}_i$
\STATE \textbf{else}
\STATE \hspace{0.5cm} Select $\mathcal{N}^\text{tree}_i$ by tree search
\STATE \hspace{0.5cm} Select $\mathcal{N}^\text{sense}_i$ with the neighboring map
\STATE \hspace{0.5cm} Compute $\eta_i$ with $\mathcal{N}^\text{sense}_i$ by Eq.~(\ref{eqn:eta})
\STATE \hspace{0.5cm} \textbf{if} $\eta_i > 1$ \textbf{then}
\STATE \hspace{1.0cm} $\mathcal{N}_i \gets \mathcal{N}^\text{tree}_i \cup \mathcal{N}^\text{sense}_i$
\STATE \hspace{0.5cm} \textbf{else}
\STATE \hspace{1.0cm} Generate $\mathcal{N}^\text{virtual}_i$ using $\mathcal{N}^\text{sense}_i$ and $\mathcal{N}^\text{robot}_i$
\STATE \hspace{1.0cm} $\mathcal{N}_i \gets \mathcal{N}^\text{virtual}_i$
\STATE \hspace{0.5cm} \textbf{end}
\STATE \textbf{end}
\STATE Compute $\mathbf{v}^\text{form}_i$ with $\mathcal{N}_i$ by Eq.~(\ref{eqn:v_form})
\end{algorithmic}
\end{algorithm}

The weight $\omega_n$ in Eq.~(\ref{eqn:v_form}) is defined as $s_n / \lVert \mathbf{p}_n - \mathbf{p}_i \rVert$ for $\eta_i > 1$, where $s_n$ denotes the area or volume of candidate node $n$. The factor $s_n$ is included in this regime because the candidate set consists of leaf nodes of varying sizes. In contrast, when $\eta_i \leq 1$, all virtual nodes are of equal size, and thus the weight depends solely on the distance. Here, $\omega_n$ is computed using a function $\phi(\lVert \mathbf{p}_n - \mathbf{p}_i \rVert / r_\text{sense})$, where $\phi(x)$ is defined as
\begin{equation*}
\phi(x) = 
\begin{cases} 
1, & x \le 0, \\ 
\frac{1}{2}\left[1+\cos(\pi x)\right], & x \in (0,1), \\
0, & x \ge 1.
\end{cases}
\end{equation*}
The overall procedure for computing the forming velocity of robot $i$ is summarized in Algorithm~\ref{alg:v_form}.

Compared with \cite{sun2023mean}, although our controller shares a mean-shift-style weighted average form, it is designed to accommodate the hierarchical and heterogeneous tree representation. First, shape entering is achieved via an explicit local tree search rather than an artificial grayscale gradient field, and the entering behavior is integrated into the mean-shift-based velocity instead of using a separate entering term. Second, neighborhood sensing relies on an auxiliary neighboring map, since direct adjacent cell enumeration is infeasible on a tree structure. Third, within-shape exploration is computed on temporary virtual nodes generated from the neighborhood sensing results rather than on uniform grid cells. This design allows the controller to operate directly on the compressed tree map while preserving the memory efficiency.

\begin{figure*}[!t]
\centering
\includegraphics[width=1.0\textwidth]{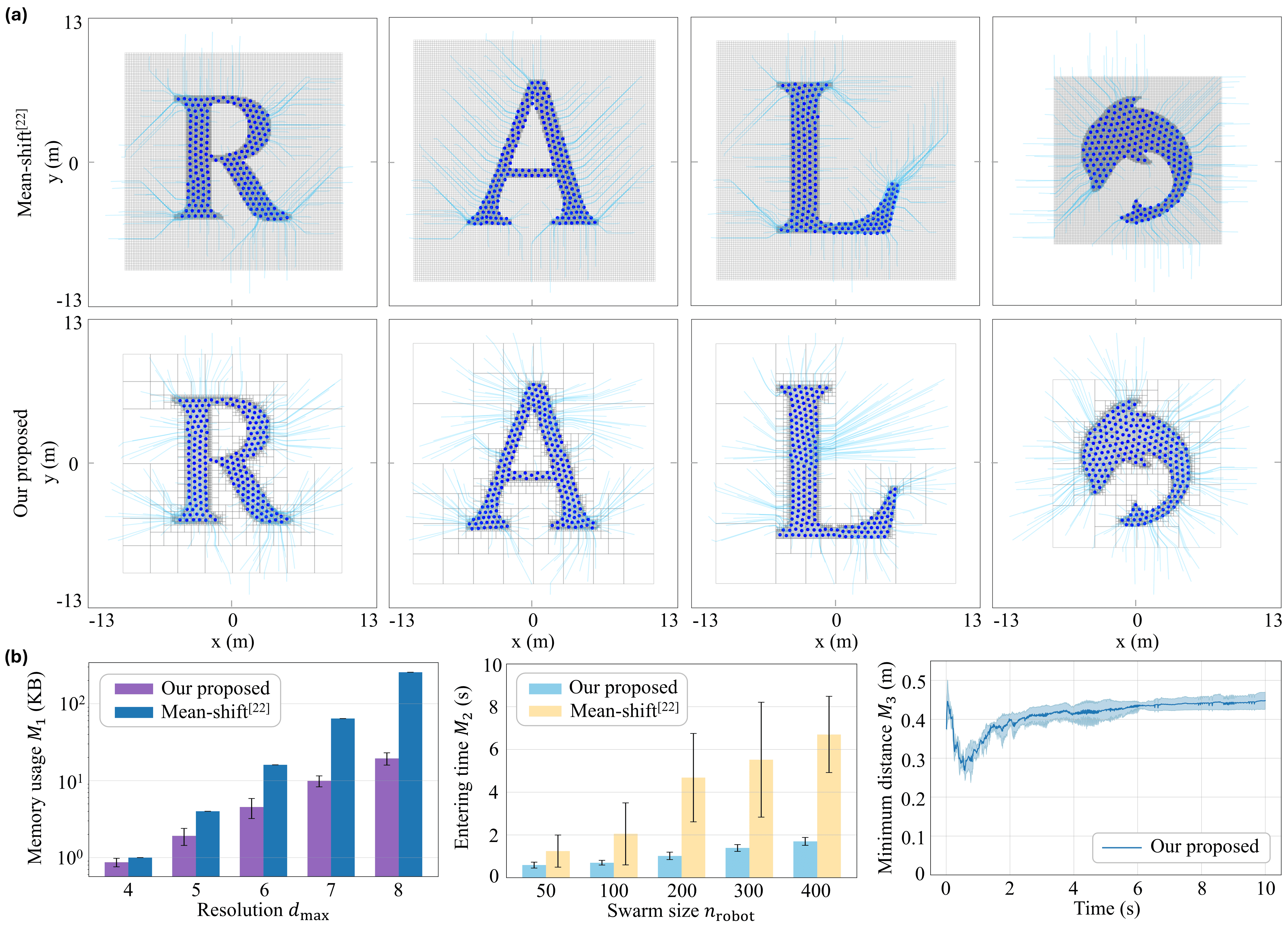}
\caption{2D Simulation Results. (a) Robot trajectories for four target shapes with $n_\text{robot}=200$. The first row shows baseline results~\cite{sun2023mean} using full-grid maps, and the second row shows results of the proposed method with tree maps. (b) Evaluation metrics over four target shapes: memory usage versus $d_\text{max}$ from 4 to 8, entering time versus $n_\text{robot}$ from 50 to 400, and the minimum distance with $n_\text{robot}=200$.}
\label{fig:2D}
\end{figure*}

\subsubsection{Collision-Avoidance Velocity}
This velocity command is designed to prevent collisions between robots. The collision-avoidance velocity $\mathbf{v}^\text{avoid}_{i}$ for robot $i$ is defined as
\begin{equation}
\label{eqn:v_avoid}
\mathbf{v}^\text{avoid}_{i} =  
\sum_{j\in \mathcal{N}^\text{robot}_i}{
\kappa_2 \mu \left(
\frac{\left\|\mathbf{p}_i - \mathbf{p}_j\right\|}{(1+\alpha)r_\text{avoid}}
\right)
(\mathbf{p}_i - \mathbf{p}_j)
},
\end{equation}
where $\mathbf{p}_j$ is the position of a neighboring robot $j \in \mathcal{N}^\text{robot}_i$. Since UAVs are used in our experiments, the downwash effect must be considered. Downwash refers to the downward airflow generated by UAV propellers, which can destabilize UAVs flying beneath them. To mitigate this effect, we set the collision-avoidance distance in the vertical direction larger than that in the horizontal directions by introducing a safe ratio $\alpha$, where $\alpha=0$ in the $x$- and $y$-directions and $\alpha>0$ in the $z$-direction. The constant $\kappa_2$ is a positive control gain and the function $\mu(s)$ is defined as $\frac{1}{s} - 1$ for $s \le 1$ and $0$ otherwise. An example is shown in Fig.~\ref{fig:overview}(c), where robots 4, 10, and 11 are repelled by each other.

\section{Experiments}
\label{sec:exper}
To evaluate the performance of the proposed shape assembly scheme, both 2D and 3D simulations and physical experiments are conducted, with results shown and analyzed in this section.

\subsection{2D Simulation Results}
In the 2D simulation, the proposed scheme is compared with~\cite{sun2023mean} under identical experimental settings. Robot parameters are set as $r_\text{avoid} = 0.6\,\text{m}$, $r_\text{sense} = 1.5\,\text{m}$, and $v_\text{max} = 10\,\text{m/s}$, while the control parameters are $\kappa_1 = 20$, $\kappa_2 = 25$. In the baseline, $\kappa_1$ governs both entering and exploration, while $\kappa_2$ handles collision avoidance (see~\cite{sun2023mean} for more details). Both methods use the same resolution: $d_\text{max} = 7$ for the tree map and $n_\text{cell} = 128 \times 128$ for the full-grid map. Both maps are globally decoded at the simulation domain center, with an identical randomly generated initial robot distribution. Simulations run for 1000 time steps with a frequency of $100\,\text{Hz}$. We evaluate four target shapes, including three letters (``R'', ``A'', ``L'') and one geometry (``Dolphin'').

\begin{figure*}[!t]
\centering
\includegraphics[width=1.0\textwidth]{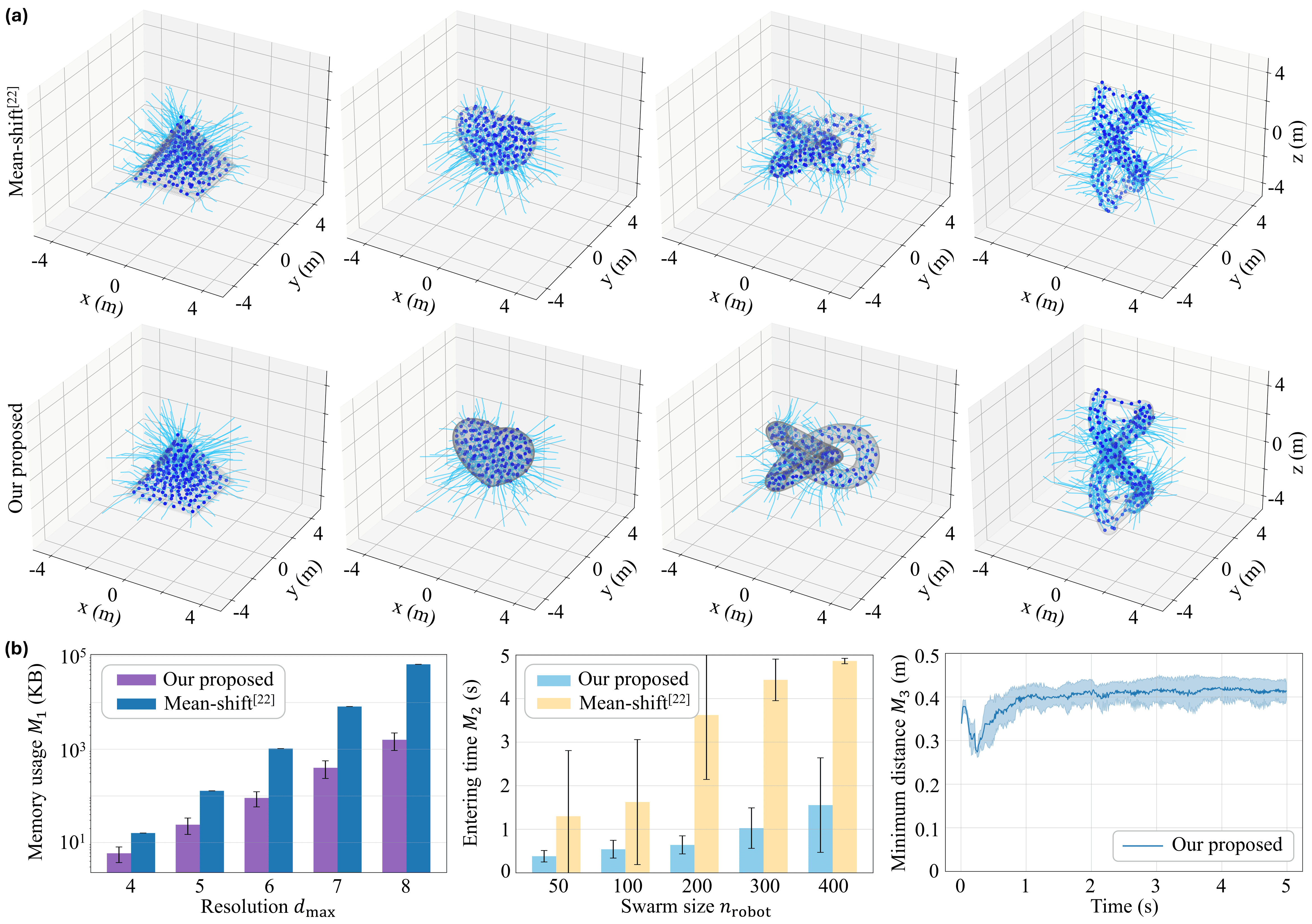}
\caption{3D Simulation Results. (a) Robot trajectories for four target shapes with $n_\text{robot}=200$: baseline method~\cite{sun2023mean} (first row) and proposed method (second row). For clarity, the target shapes are shown only as gray shells. (b) Evaluation metrics over four target shapes: memory usage versus $d_\text{max}$ (4 to 8), entering time versus $n_\text{robot}$ (50 to 400), and the minimum distance with $n_\text{robot}=200$.}
\label{fig:3D}
\end{figure*}

Throughout this section, we evaluate performance using three quantitative metrics. First, memory usage $M_1$ measures the memory required to encode the target shapes, which is defined in Eqs.~(\ref{eqn:memory}) and (\ref{eqn:memory-cmp}) for the proposed and baseline methods, respectively. Second, entering time $M_2$ is defined as the time required for all robots to enter the target shape. Third, minimum distance $M_3$ assesses the collision avoidance performance; it is computed as the minimum inter-robot distance within the swarm, so collisions are avoided if $M_3 > 0$.

Fig.~\ref{fig:2D}(a) shows the assembled shapes and robot trajectories for the baseline (mean-shift) method and our method with $n_\text{robot} = 200$. Both algorithms successfully guide the swarm to assemble the desired shapes. To compare performance systematically, Fig.~\ref{fig:2D}(b) reports three metrics over four trials (one per shape). For $M_1$, $d_\text{max}$ ranges from 4 to 8; for $M_2$, $n_\text{robot}$ increases from 50 to 400; and for $M_3$, $n_\text{robot}$ is fixed at 200. In all cases, results are computed over the four trials. For $M_3$, the solid line shows the mean and the shaded area denotes the standard deviation across trials. From these results, the tree map reduces memory usage by a factor of 2 to 16 compared with the full-grid map, with the gap increasing with resolution. Despite this reduction, the proposed method improves entering efficiency, reducing entering time by a factor of two to three, with the benefit increasing with swarm size. Moreover, the minimum distance remains strictly positive throughout the simulation, indicating collision-free motion. This improvement is mainly due to the controller design: the baseline follows local grayscale gradients, which can yield longer trajectories and may trap robots in gradient valleys near sharp shape features, causing clustering and queuing. In contrast, our approach uses tree search with a neighboring map to provide a more extended spatial view, guiding robots toward target regions more directly with minimal computational overhead (see Subsection II-B for details).

\subsection{3D Simulation Results}
Both the baseline and the proposed scheme are further validated in 3D simulations. Compared with 2D simulations, the experimental setting adjustments include parameters $r_\text{avoid} = 0.5\,\text{m}$, $r_\text{sense} = 0.8\,\text{m}$, $\alpha=0.5$ in the z-direction, and resolution $d_\text{max} = 6$. All simulations are run for 500 time steps. We select four 3D target shapes: ``Pyramid'', ``Heart'', ``Knot'', and ``DNA''. Fig.~\ref{fig:3D}(a) shows the final assembled shapes using the baseline and the proposed algorithm with $n_\text{robot} = 200$ in a 3D view. These results demonstrate that both algorithms can successfully enable a large robot swarm to assemble 3D target shapes, even for the sophisticated DNA shape.

\begin{figure*}[!t]
\centering
\includegraphics[width=1.0\textwidth]{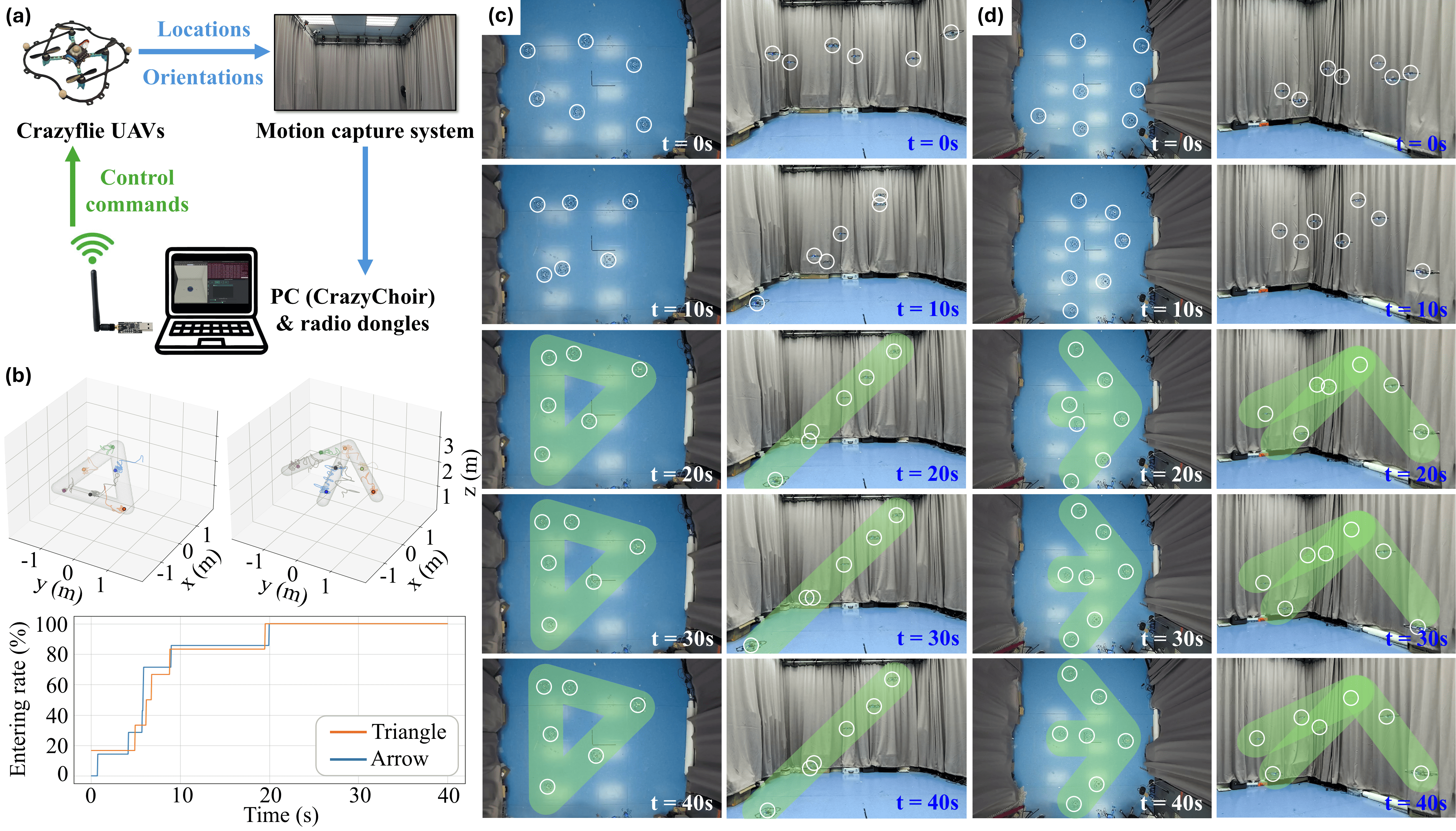}
\caption{Physical Experiment Results. (a) Experimental system setup. (b) Robot trajectories and entering rate. Snapshots of the swarm during the experiments of (c) triangle assembly and (d) arrow assembly are shown from two viewing angles.}
\label{fig:exp}
\end{figure*}

Fig.~\ref{fig:3D}(b) presents the performance metrics, evaluated in the same setting as in 2D. For memory usage, the tree map encoding achieves up to 65 times smaller memory at $d_\text{max}=8$, representing a more substantial improvement than in 2D. For entering time, the shape entry efficiency gain is also more pronounced in 3D, reaching a factor of approximately three to four. Moreover, as $n_\text{robot}$ increases from 50 to 400, the entering time for the proposed method grows only from about 40 to 160 time steps, indicating an approximately linear scaling. Notably, the entering times in 3D closely match those in 2D for the same swarm sizes, albeit with larger variation across shapes, suggesting robust and largely dimension-independent performance. Finally, the minimum distance data again indicates reliable collision-avoidance behavior.

\subsection{Physical Experiment Results}
In the physical experiment, a 3D triangle and a 3D arrow are assembled by 6 and 7 Crazyflie UAVs, respectively, within a $4\,\text{m} \times 5\,\text{m} \times 3\,\text{m}$ space, as shown in Fig.~\ref{fig:exp}. The experimental parameters are consistent with those used in the 3D simulations, except for $r_\text{sense} = 1.2\,\text{m}$, $v_\text{max} = 0.3\,\text{m/s}$, and $d_\text{max} = 5$. The system configuration is illustrated in Fig.~\ref{fig:exp}(a), consisting of UAVs, a PC running an open-source aerial swarm control toolbox (CrazyChoir)~\cite{pichierri2023crazychoir}, and a motion capture system. The motion capture system and the ground station PC are used as an engineering implementation to ensure reliable state acquisition and to address onboard computation constraints. They do not alter the distributed information flow: the PC runs independent per-robot control threads in parallel; each thread uses only that UAV's current state and neighborhood information and transmits the control command to that UAV at a control frequency of $20\,\text{Hz}$. This implementation practice is consistent with the baseline method~\cite{sun2023mean}.

During the experiments, all UAVs first take off and hover at approximately $1\,\text{m}$, then the control algorithm is activated and executed for $40\,\text{s}$. Note that the times reported in Fig.~\ref{fig:exp} refer only to the control phase, excluding take-off and hovering. Fig.~\ref{fig:exp}(b) shows the recorded trajectories of all robots and the entering rate for the two shapes. The entering rate denotes the percentage of robots inside the shape. In both experiments, all robots enter the shape after approximately $20\,\text{s}$. The corresponding snapshots from the top and oblique side views are presented in Fig.~\ref{fig:exp}(c) and (d). The snapshots show that both shapes are assembled within $30\,\text{s}$ and maintained until $40\,\text{s}$. The discrepancy between the experimental and simulation assembly efficiency arises mainly from differences in maximum velocity constraints and control frequency due to hardware limitations, as well as real-world system dynamics such as downwash and ground effects. These results demonstrate that the UAV swarm can assemble different target shapes efficiently and robustly, thereby validating the real-world applicability of the proposed algorithm. Note that these experiments are intended as a feasibility validation under practical real-world constraints, whereas the primary evidence for scalability is provided by the large-scale simulations.

\section{Conclusion}
\label{sec:concl}
This letter has presented a novel framework for robot swarm shape assembly in both 2D and 3D scenarios. The proposed framework integrated a memory-efficient, tree-based shape encoding-decoding scheme with a behavior-based, distributed assembly controller. Compared with a representative image-based baseline, our approach reduced memory usage by one to two orders of magnitude, which is critical for fully distributed systems and resource-constrained robots. Furthermore, 2D and 3D simulations with up to 400 robots demonstrated dimension-independent performance of our method, achieving two to four times faster shape entry compared with the baseline. Finally, experiments with a swarm of 6 to 7 UAVs validated the real-world feasibility of our framework. Future research will focus on eliminating the global frame assumption, mitigating the downwash effect, reducing dependency on the motion capture system, and developing robust communication strategies to cope with intermittent links in 3D deployments.

\bibliographystyle{IEEEtran}
\bibliography{references}

\end{document}